# Specifying a shallow grammatical representation for parsing purposes

Atro Voutilainen and Timo Järvinen
Research Unit for Multilingual Language Technology
P.O. Box 4
FIN-00014 University of Helsinki
Finland
Atro.Voutilainen,Timo.Jarvinen@Helsinki.FI

**Abstract**

Is it possible to specify a grammatical representation (descriptors and their application guidelines) to such a degree that it can be consistently applied by different grammarians e.g. for producing a benchmark corpus for parser evaluation? Arguments for and against have been given, but very little empirical evidence. In this article[1] we report on a double-blind experiment with a surface-oriented morphosyntactic grammatical representation used in a large-scale English parser. We argue that a consistently applicable representation for morphology and also shallow syntax can be specified. A grammatical representation with a near-100% coverage of running text can be specified with a reasonable effort, especially if the representation is based on structural distinctions (i.e. it is structurally resolvable).

## 1 Introduction

The central task of a parser is to assign grammatical descriptions onto input sentences. Evaluating a parser's output (as well as designing a computational lexicon and grammar) presupposes a predefined, parser-independent specification of the grammatical representation.

Perhaps surprisingly, the possibility of specifying a workable grammatical representation is a matter of controversy, even at lower levels of analysis, e.g. morphology (incl. parts of speech).

---

[1] This paper is published in *Proceedings of the Seventh Conference of the European Chapter of the Association for Computational Linguistics*, Dublin, 1995.

Consider the following setting (the double-blind experiment). Two linguists trained to apply a tag set to running text according to application guidelines (a "style sheet") are to analyse a given data individually. The results are then automatically compared, and the differences are jointly examined by these linguists to see whether the differences are due to inattention, or whether they are intentional (i.e. there is a genuine difference in analysis). – How many percentage points of all words in running text are retain a different analysis after the differences due to inattention have been omitted? The higher this percentage, the more susceptible seems the possibility of specifying a workable grammatical representation.

According to a pessimistic view (e.g. Church 1992), the part of speech of several percentage points of words in running text is impossible to agree on by different judges, even after negotiations. A more optimistic view can be found in (Leech and Eyes 1993, p. 39; Marcus *et al.* 1993, p. 328); they argue that a near-100% interjudge agreement is possible, provided the part-of-speech annotation is done carefully by experts. Unfortunately, they give very little empirical evidence for their position e.g. in terms of double-blind experiments.

Supposing defining these lower levels of grammatical representation is so problematic, the more distinctive levels should be even more difficult. If specifying the task of the parser – what the parser is supposed to do – turns out to be so problematic, one could even question the rationality of natural language parser design as a whole. In other words, the controversy regarding the specifiability of a grammatical representation is a fundamental issue.

In this article we report on a double-blind experiment with a surface-oriented morphosyntactic grammatical representation used by a large-scale English parser. We show that defining a grammatical representation is possible, even relatively straightforward. We present results from part-of-speech annotation and shallow syntactic analysis. Our three main findings are:

1. A practically 100% interjudge agreement can be reached at the level of morphological (incl. part-of-speech) analysis provided that (i) the grammatical representation is based on structural distinctions and (ii) the individual descriptive practices of the most frequent 'problem cases' are properly documented.

2. A shallow dependency-oriented functional syntax can be defined, very much like a morphological representation. The only substantial difference seems to be that somewhat more effort for documenting the individual solution is needed at the level of syntax.

3. A grammatical representation (morphosyntactic descriptors and their application guidelines) can be specified with a reasonable effort. In addition to general descriptive principles, only a few dozen construction-specific entries seem necessary for reaching a high coverage of running text.

In short: In this paper we give empirical evidence for the possibility of specifying a grammatical representation in enough detail to make it (almost) consistently applicable. What we are less specific about here is the exact formal properties that make a representation easy to specify; this topic remains open for future investigation.

## 2  Grammatical representation in English Constraint Grammar

In the experiment to be reported in Section 3, we employed the grammatical representation that defines the descriptive task of the English Constraint Grammar Parser ENGCG (Karlsson *et al.* (eds.) 1995).[2]

### 2.1  Morphology

The morpholexical component in ENGCG employs 139 morphological tags for part of speech, inflection, derivation and certain syntactic properties (e.g. verb classification). Each morphological analysis usually consists of several tags, and many words get several analyses as alternatives. The following analysis of the sentence *That round table might collapse* is a rather extreme example:

```
"<*that>"
    "that" <*> <**CLB> CS
    "that" <*> DET CENTRAL DEM SG
    "that" <*> ADV
    "that" <*> PRON DEM SG
    "that" <*> <**CLB> <Rel> PRON SG/PL
"<round>"
    "round" <SVO> V SUBJUNCTIVE VFIN
    "round" <SVO> V IMP VFIN
    "round" <SVO> V INF
    "round" <SVO> V PRES -SG3 VFIN
    "round" PREP
    "round" N NOM SG
    "round" A ABS
    "round" ADV
"<table>"
    "table" N NOM SG
```

---

[2] A list of the ENGCG tags can be retrieved via e-mail by sending an empty mail message to engcg-info@ling.helsinki.fi. The returned document will also tell how to analyse own samples using the ENGCG server.

```
    "table" <SVO> V SUBJUNCTIVE VFIN
    "table" <SVO> V IMP VFIN
    "table" <SVO> V INF
    "table" <SVO> V PRES -SG3 VFIN
"<might>"
    "might" N NOM SG
    "might" V AUXMOD VFIN
"<collapse>"
    "collapse" N NOM SG
    "collapse" <SVO> V SUBJUNCTIVE VFIN
    "collapse" <SVO> V IMP VFIN
    "collapse" <SVO> V INF
    "collapse" <SVO> V PRES -SG3 VFIN
"<$.>"
```

The morphological analyser produces about 180 different tag combinations. To compare the ENGCG morphological description with another well-known tag set, the Brown Corpus tag set: ENGCG is more distinctive in that the part of speech distinction is spelled out in the description of determiner–pronoun, preposition–conjunction, and determiner–adverb–pronoun homographs, as well as uninflected verb forms, which are represented as ambiguous due to the subjunctive, imperative, infinitive and present tense readings. On the other hand, ENGCG does not spell out part-of-speech ambiguity in the description of *-ing* and nonfinite *-ed* forms, noun–adjective homographs when the core meanings of the adjective and noun readings are similar, nor abbreviations vs. proper vs. common nouns. Generally, the ENGCG morphological tag set avoids the introduction of structurally unjustified distinctions.

## 2.2 Syntax

ENGCG syntax employs 30 dependency-oriented functional tags that indicate the surface-syntactic roles of nominal heads (subject, object, preposition complement, apposition, etc.) and modifiers (premodifiers, postmodifiers). The shallow structure of verb chains is also given – the tag set distinguishes between auxiliaries and main verbs, finite and nonfinite. Also the structure of adverbials as well as prepositional and adjective phrases is given, though some of the attachments of adverbials is left underspecified.

Finally, a disambiguated sample analysis of the above sample sentence:

```
"<*that>"
    "that" <*> DET CENTRAL DEM SG @DN>
"<round>"
```

```
    "round" A ABS @AN>
"<table>"
    "table" N NOM SG @SUBJ
"<might>"
    "might" V AUXMOD VFIN @+FAUXV
"<collapse>"
    "collapse" <SVO> V INF @-FMAINV
"<$.>"
```

Syntactic tags are flanked with the @-sign;[3] morphological tags and the base form are given to the left of the syntactic tags.

# 3 The experiment

This section reports on an experiment on part-of-speech and syntactic disambiguation by human experts (the authors of this article). Three 2,000-word texts were successively used: a software manual, a scientific magazine, and a newspaper.

## 3.1 Setting

The experiment was conducted as follows.

1. The text was morphologically analysed using the ENGCG morphological analyser. For the analysis of unrecognised words, we used a rule-based heuristic component that assigns morphological analyses, one or more, to each word not represented in the lexicon of the system.

2. Two experts in the ENGCG grammatical representation independently marked the correct alternative analyses in the ambiguous input, using mainly structural, but in some structurally unresolvable cases also higher-level, information. The corpora consisted of continuous text rather than isolated sentences; this made the use of textual knowledge possible in the selection of the correct alternative. In the rare cases where two analyses were regarded as equally legitimate, both could be marked. The judges were encouraged to consult the documentation of the grammatical representation.

---

[3] "@DN>" represents determiners; "@AN>" represents premodifying adjectives; "@SUBJ" represents subjects; "@+FAUXV" represents finite auxiliaries; and "@-FMAINV" represents nonfinite main verbs.

3. These tagged versions were compared to each other using the Unix *sdiff* program.

4. The differences were jointly examined by the judges in order to see whether they were due to (i) inattention, (ii) incomplete specification of the grammatical representation or (iii) an undecidable analysis.

5. A 'consensus' version of the tagged corpus was prepared. Usually only a unique analysis was given. However, there were three situations where a multiple analysis was accepted:

    - When the judges disagree about the correct analysis even after negotiations. In this case, comments were added to distinguish it from the other two types.
    - Neutralisation: both analyses were regarded as equivalent. (This often indicates a redundancy in the lexicon.)
    - Global ambiguity: the sentence was agreed to be globally ambiguous.

6. Whenever an undefined construction was detected during the joint examination, the grammar definition manual was updated.

7. The preparation of the syntactic version was the next main step. For each contextually appropriate morphological reading, all syntactic tags were introduced with a mapping program. An example:[4]

```
"<*that>"
    "that" <*> DET CENTRAL DEM SG @DN>
"<round>"
    "round" A ABS @AN>
"<table>"
    "table" N NOM SG @NPHR @SUBJ @OBJ
            @I-OBJ @PCOMPL-S @PCOMPL-O
            @APP @NN> @<P @O-ADVL
"<might>"
    "might" V AUXMOD VFIN @+FAUXV
"<collapse>"
    "collapse" <SVO> V INF @-FMAINV
                @<P-FMAINV @<NOM-FMAINV
"<$.>"
```

---

[4] "@NPHR" represents stray nominal heads; "@OBJ" represents objects; "@I-OBJ" represents indirect objects; "@PCOMPL-S" represents subject complements; "@PCOMPL-O" represents object complements; "@APP" represents appositions; "@NN>" represents premodifying nouns (and nonfinal noun parts in compounds); "@<P" represents nominal preposition complements; "@O-ADVL" represents nominal adverbials; "@<P-FMAINV" represents nonfinite main verbs as preposition complements; and "@<NOM-FMAINV" represents postmodifying nonfinite main verbs.

| text | news | technical | magazine | total |
|---|---|---|---|---|
| words | 1999 | 1999 | 2073 | 6071 |
| morph.tags/word in input | 1.78 | 1.95 | 1.72 | 1.82 |
| morphologically ambiguous words | 41.1% | 45.4% | 36.8% | 41.0% |
| agreement after mechanical comparison | 99.3% | 99.3% | 99.1% | 99.2% |
| updates to morphology manual | 1 | 1 | 1 | 3 |
| agreement after negotiations | 100% | 100% | 100% | 100% |
| morph.tags/word in consensus corpus | 1.00 (+2) | 1.00 (+0) | 1.00 (+1) | 1.00 (+3) |
| syn.tags/word in input | 3.50 | 3.53 | 3.36 | 3.46 |
| syntactically ambiguous words | 42.0% | 41.9% | 44.9% | 42.9% |
| agreement after mechanical comparison | 95.8% | 97.0% | 97.4% | 96.8% |
| updates to syntax manual | 5 | 1 | 1 | 7 |
| agreement after negotiations | 100% | 100% | 100% | 100% |
| syn.tags/word in consensus corpus | 1.01 (+18) | 1.01 (+11) | 1.00 (+3) | 1.01 (+32) |

Figure 1: Results of a tagging test.

8. Steps 2–6 were applied to these syntactic ambiguities.

This procedure was successively applied to the three texts to see how much previous updates of the grammar definition manual decreased the need for further updates and how much the interjudge agreement might increase even after the first mechanical comparison (cf. Step 3).

## 3.2 Results

The results are given in Figure 1.

Some comments are in order, first about morphology.

- The initial consistency rate was constantly above 99%.

- After negotiations, the judges agreed about the correct analysis or analyses in all cases. The vast majority of the initial differences were due to inattention, and the remaining few to incomplete specification of the morphological representation. Some representative examples about these jointly examined differences are in order. (Words followed by an expression of the form **(X/Y)** were initially tagged differently by the judges. After joint examination, **Y** was agreed to be the correct alternative in all cases but (5), where **X** and **Y** were regarded as equally possible.)[5]

---

[5]Before an "of" phrase, the pronoun/numeral distinction of "one" was regarded as neutralised. This observation was also added to the morphology manual.

1. As we go(V INF / V PRES) to(INFMARK / PREP) press(V INF / N), George Bush's decision not to sign the Biodiversity Convention, and Britain's apparent intention to follow suit, are grievous blows..
2. .. they were circulating a letter expressing concern that(PRON REL / CS) it would give the developing countries a blank cheque to demand money from donors to finance sustainable development.
3. That(PRON DEM / CS) there was no outburst of protest over the new policy suggests that public anxiety over genetic engineering has ebbed in recent years.
4. The value-added information is the kind(A / N) we want ourselves."
5. .. they had not seen before at one(NUM / PRON) of the busiest times of the school year.
6. I don't think people get(V INF / V PRES) a great deal from bald figures.
7. She had to ask because some of the six-year-olds from other schools who attend(V INF / V PRES) her classes know the names of as(PREP / AD-A>) many hard drugs as she does.

- Only three updates were needed to the morphological part of the manual.
- Though multiple analyses were considered acceptable in the case of (even semantically) undecidable situations, very few were actually needed: only 3 words out of 6,071 received two analyses (for example, it was agreed that *more* could be analysed both as an adverb and as a pronoun in *.. free trade will mean you destroy more.*).

Next, some observations about syntax.

- At the level of syntax, most of the initial differences were identified as obvious mistakes, e.g.:
  - *He was (@+FMAINV / @+FAUXV) addressing his hosts ..*
- Sometimes, however, there was a need to discuss the descriptive policies. Consider the following sentence fragment:[6]
  - *.. that managers'(@GN>) keeping (@-FMAINV / @SUBJ) in(@ADVL / @<NOM) touch with employees enhances communication ..*

---

[6] "@GN>" represents genitival premodifiers, and "@<NOM" represents postmodifiers.

In principle, *managers'* could be described as a subject in a nonfinite clause, and *keeping* accordingly as a nonfinite main verb. However, the ENGCG syntactic representation does not recognise the subject category in nonfinite clauses; therefore, in the name of consistency, *keeping* in the above example should be assigned a nominal rather than a verbal function – finite clause subject, in this case.

- Initially, the syntactic representation was less completely specified than the morphological representation. The grammar definition manual initially comprised twelve entries for syntactic functions; seven additions were made during the experiment. This had a positive effect: the initial disagreements decreased from 4.2% to 2.6% during the three rounds.

- The entries in the syntax manual can be classified into three types:

  1. Two or more alternatives are structurally plausible, but one is to be consistently preferred; e.g. *A number that occurs after a proper noun and is surrounded by commas is a postmodifier (rather than an apposition)*
  2. Elimination of a distinction in certain contexts; e.g. *Premodifying -ing forms are to be analysed as adjectives (rather than as nouns)*
  3. An unorthodox policy is adopted; e.g. *In sentences with a formal subject, what is usually regarded as a notional subject is here analysed as a subject complement.*

- Multiple analyses were given to 32 words (0.5% of all words). PP attachment, in particular the distinction between clause level (@ADVL) and postmodifying (@<NOM) functions, proved to be the most difficult syntactic phenomenon to define uniquely; often the analyses remained somewhat indeterminate. With the first sample, 5.7% of all prepositions were initially annotated differently; even with the last sample, 4.5% of the initial analyses differed. Unsurprisingly, a frequent agreement in the analysis of these cases was to accept both alternatives as legitimate.

- Some further examples are in order. These examples show some possible structural ambiguities in which text-level semantic information was needed to decide upon the preferred analysis **Y** over less plausible alternative **X**. Note that the adopted decision often determines the correct analysis of one or more subsequent items (the "domino effect").

  1. *.. his priority was (@+FAUXV / @+FMAINV) keeping his country's biotechnology industry free ..*
  2. *Germany wants the heads of European governments and perhaps Japan (@OBJ / @<P) to issue a 'declaration of like-minded parties".*

3. *We were (@-FMAINV / @+FAUXV) pleased (@PCOMPL-S / @-FMAINV) with (@<NOM / @ADVL) the report.*

The last type was recurrent because the ENGCG morphology offers only a past participle reading to *-ed*-forms. We prefer the verbal reading and predicative *-ed*-forms are listed as exceptions in the coding manual.

## 4 Conclusion

A satisfactory definition of the grammatical representation appears possible, not only at the level of morphology, but also at the level of shallow dependency-oriented functional syntax. In our experiments, a practically 100% consensus was reached at both these levels during the joint examination. Our results agree, at least at the level of morphology, with (Leech and Eyes 1993; Marcus *et al.* 1993). In our experiment, the main differences between morphology and syntax were that (i) specifying the syntactic representation takes a few more pages in the definition manual, and (ii) there seem to be more cases in syntax where multiple analyses have to be accepted – but relatively few even then.

The grammatical representation should employ intuitively clear grammatical descriptors that (i) represent all constructions in the language and (ii) reflect distributional distinctions. Proposing a too fine-grained classification of e.g. *-ing* forms, as may be the case in the tagged Brown Corpus, can make the principled selection of the appropriate analysis very difficult, even with detailed manuals.

As a minor point we may add that errors due to inattention tend to occur in the preparation of e.g. benchmark corpora; however, almost all of them can be eliminated using the double-blind method.

## Acknowledgements

We would like to thank Jussi Piitulainen, Pasi Tapanainen and two EACL referees for useful comments on a previous version of this paper. The usual disclaimers hold.